\documentclass[10pt,journal]{IEEEtran}

\ifCLASSOPTIONcompsoc
\usepackage[nocompress]{cite}
\else
\usepackage{cite}
\fi
\usepackage{cite}
\usepackage{hyperref}
\usepackage{url}

\usepackage[utf8]{inputenc} 
\usepackage[T1]{fontenc}    
\usepackage{hyperref}       
\usepackage{url}            
\usepackage{booktabs}       
\usepackage{amsfonts}       
\usepackage{nicefrac}       
\usepackage{microtype}      
\usepackage{epsfig}
\usepackage{graphicx}
\usepackage{amsmath}
\usepackage{amssymb}
\usepackage{epstopdf}
\usepackage{color,soul}

\begin{document}

	\title{Deep Learning with Darwin: Evolutionary Synthesis of Deep Neural Networks}

	\author{Mohammad Javad  Shafiee,~\IEEEmembership{Student Member,~IEEE,}
		 Akshaya Mishra,~\IEEEmembership{Member,~IEEE,}
		and~Alexander Wong,~\IEEEmembership{Senior Member,~IEEE}
		\IEEEcompsocitemizethanks{
			\IEEEcompsocthanksitem M.~J. Shafiee, A. Mishra and A. Wong are with the Department
			of  Systems Design Engineering, university of Waterloo, Waterloo,
			ON, Canada.\protect\\
			E-mail: mjshafiee@uwaterloo.ca
		}
		\thanks{Manuscript received XXXX, 2016; revised XXXX, 2016.
			
			This research has been supported by Canada Research Chairs programs, Natural Sciences and Engineering Research Council of Canada (NSERC), and the Ministry of Research and Innovation of Ontario.  The authors also thank Nvidia for the GPU hardware used in this study through the Nvidia Hardware Grant Program.
			
	}}

	\markboth{IEEE SIGNAL PROCESSING LETTERS,~Vol.~X, No.~X, XXXX~2016}%
	{Shell \MakeLowercase{\textit{et al.}}: Bare Demo of IEEEtran.cls for Computer Society Journals}
	
	\maketitle

\begin{abstract}
Taking inspiration from biological evolution, we explore the idea of ``\textit{Can deep neural networks \textbf{evolve} naturally over successive generations into highly efficient deep neural networks?}'' by introducing the notion of synthesizing new highly efficient, yet powerful deep neural networks over successive generations via an evolutionary process from ancestor deep neural networks.  The architectural traits of ancestor deep neural networks are encoded using synaptic probability models, which can be viewed as the `DNA' of these networks.  New descendant networks with differing network architectures are synthesized based on these synaptic probability models from the ancestor networks and computational environmental factor models, in a random manner to mimic heredity, natural selection, and random mutation.  These offspring networks are then trained into fully functional networks, like one would train a newborn, and have more efficient, more diverse network architectures than their ancestor networks, while achieving powerful modeling capabilities.  Experimental results for the task of visual saliency demonstrated that the synthesized `evolved' offspring networks can achieve state-of-the-art performance while having network architectures that are significantly more efficient (with a staggering $\sim$48-fold decrease in synapses by the fourth generation) compared to the original ancestor network.
\end{abstract}

\section{Introduction}
\vspace{-0.1 cm}
\IEEEPARstart{D}{eep}
 learning, especially deep neural networks~\cite{lecun2015deep,graves2013speech,bengio2009learning,tompson2014joint} have shown considerable promise through tremendous results in recent years, significantly improving the accuracy of a variety of challenging problems when compared to other machine learning methods~\cite{krizhevsky2012imagenet,farabet2013learning,simonyan2014very,hinton2012deep,hannun2014deep,amodei2015deep}.  However, deep neural networks require high performance computing systems due to the tremendous quantity of computational layers they possess, leading to a massive quantity of parameters to learn and compute.  This issue of architectural complexity has increased greatly in recent years~\cite{simonyan2014very,srivastava2015training,szegedy2015going}, driven by the demand for increasingly deeper and larger deep neural networks to boost modeling accuracy.  As such, it has become increasingly more difficult to take advantage of such complex deep neural networks in scenarios where computational and energy resources are scarce.

To enable the widespread use of deep learning, there has been a recent drive towards obtaining highly-efficient deep neural networks with strong modeling power.  Much of the work in obtaining efficient deep neural networks have focused on deterministically compressing trained deep neural networks~\cite{lecun1989optimal}, using traditional lossless and lossy compression techniques such as quantization~\cite{gong2014compressing,han2015deep}, deterministic pruning~\cite{lecun1989optimal,han2015learning}, Huffman coding~\cite{han2015deep}, and hashing~\cite{chen2015compressing}.  Rather than attempting to take an existing deep neural network and compress it into a smaller representation heuristically, we instead consider the following idea: \textit{Can deep neural networks \textbf{evolve} naturally over successive generations into highly efficient deep neural networks?}  Using an example of evolutionary progress towards efficiency from nature, a recent study by Moran \mbox{\it et al.}~\cite{moran2015energetic} proposed that the eyeless Mexican cavefish evolved to lose its vision system over generations due to the high metabolic cost of vision.  Therefore, by evolving naturally over generations in a way where the cavefish lost its vision system, the amount of energy expended is significantly reduced and thus improves survivability in subterranean habitats where food availability is low.  The ability to mimic the biological evolutionary process for the task of producing highly-efficient deep neural networks over successive generations can have considerable benefits.

In this study, we entertain a different notion for producing highly-efficient deep neural networks by introducing the evolutionary synthesis of deep neural networks over successive generations based on ancestor deep neural networks.  While the idea of leveraging evolutionary computation concepts for training and generating deep neural networks have been previously explored in literature~\cite{Angeline,Stanley,Stanley2,Gauci,Tirumala}, there are significant key differences between these previous studies and this study:
\begin{itemize}
\item While previous studies have focused on improving the accuracy and training of deep neural networks, to the best of the authors' knowledge this study is the first to explore and focus on the notion of evolutionary synthesis of deep neural networks with high network architectural efficiency over successive generations.
\item  While the evolutionary computational approaches leveraged by these previous studies are classical approaches such as  genetic algorithms and evolutionary programming, this study introduces a new probabilistic framework where evolution mechanisms such as genetic encoding and environmental conditions are modeled via probability distributions, and the stochastic synthesis process leverages these probability models to produce deep neural networks at successive generations.  To the best of the authors' knowledge, this study is the first to leverage a probabilistic approach to evolutionary synthesis of deep neural networks.
\item To the best of the authors' knowledge, the new approach introduced in this study is the first to achieve evolution and synthesis of deep neural networks with very deep, large neural network architectures that have been demonstrated to provide great performance in recent years~\cite{simonyan2014very,srivastava2015training,szegedy2015going}.  Previous studies have focused on deep neural networks with smaller and shallower network architectures, as the approaches used in such studies are more difficult to scale to very deep, large network architectures.
\end{itemize}

\section{Methodology }

The proposed evolutionary synthesis of deep neural networks is primarily inspired by real biological evolutionary mechanisms.  In nature, traits that are passed down from generation to generation through DNA may change over successive generations due to factors such as natural selection and random mutation, giving rise to diversity and enhanced traits in later generations.  To realize the idea of evolutionary synthesis for producing deep neural networks, we introduce a number of computational constructs to mimic the following mechanisms of biological evolution: i) \textbf{Heredity}, ii) \textbf{Natural Selection}, and iii) \textbf{Random Mutation}.

\noindent \textbf{Heredity.} Here, we mimic the idea of heredity by encoding the architectural traits of deep neural networks in the form of synaptic probability models, which are used to pass down traits from generation to generation.  One can view these synaptic probability models as the `DNA' of the networks.  Let $\mathcal{H}=(\mathcal{N},S)$ denote the possible architecture of a deep neural network, with $\mathcal{N}$ denoting the set of possible neurons and $S$ denoting the set of possible synapses, with $s_{k} \in S$ denoting a synapse between two neurons $(n_i, n_j) \in \mathcal{N}$.  One can encode the architectural traits of a deep neural network as $P(\mathcal{H}_g|\mathcal{H}_{g-1})$, which denotes the conditional probability of the architecture of a network in generation $g$ (denoted by $\mathcal{H}_g$), given the architecture of its ancestor network in generation $g-1$ (denoted by $\mathcal{H}_{g-1}$).

If we were to treat areas of strong synapses in an ancestor network in generation $g$ as desirable traits to be inherited by descendant networks at generation $g$, where descendant networks have a higher probability of having similar areas of strong synapses as its ancestor network, one can instead encode the architectural traits of a deep neural network as the synaptic probability $P(S_g|\mathcal{W}_{g-1})$, where $w_{g-1,k} \in \mathcal{W}_{g-1}$ encodes the synaptic strength of each synapse $s_{g-1,k}$. Modeling $P(S_g|\mathcal{W}_{g-1})$ as an exponential distribution, with the probability of each synapse in the network assumed to be independently distributed, one arrives at
\begin{align}
P(S_{g}|\mathcal{W}_{g-1})  = \prod_{i} \exp \Big(\frac{w_{g-1,i}}{Z} - 1\Big),
\label{synaptiveprob}
\end{align}
where $Z$ is a normalization constant.

\noindent \textbf{Natural Selection and Random Mutation.} The ideas of natural selection and random mutation are mimicked through the introduction of a network synthesis process for synthesizing descendant networks, which takes into account not only the synaptic probability model encoding the architectural traits of the ancestor network, but also an environmental factor model to mimic the environmental conditions that help drive natural selection, in a random manner that drives random mutation.  More specifically, a synapse is synthesized randomly between two possible neurons in a descendant network based on $P(S_g|\mathcal{W}_{g-1})$ and an environmental factor model $\mathcal{F(\mathcal{E})}$, with the neurons in the descendant network synthesized subsequently based on the set of synthesized synapses.  As such, the architecture of a descendant network at generation $g$ can be synthesized randomly via synthesis probability $P(\mathcal{H}_g)$, which can be expressed by
\begin{align}
P(\mathcal{H}_g) = \mathcal{F(\mathcal{E})} \cdot P(S_g|\mathcal{W}_{g-1}).
\end{align}
The environmental factor model $\mathcal{F(\mathcal{E})}$ can be the combination of quantitative environmental conditions that are imposed upon the descendant networks that they must adapt to.

To have a better intuitive understanding, let us examine an illustrative example of how one can impose environmental conditions using $\mathcal{F(\mathcal{E})}$ to promote the evolution of highly efficient deep neural networks.

\noindent \textbf{Efficiency-driven Evolutionary Synthesis.} One of the main environmental factors in encouraging energy efficiency during evolution is to restrict the resources available.  For example, in a study by Moran {\it et al.}~\cite{moran2015energetic}, it was proposed that the eyeless Mexican cavefish lost its vision system over generations due to the high energetic cost of neural tissue and low food availability in subterranean habitats. Their study demonstrated that the cost of vision is about 15\% of resting metabolism for a 1-g eyed phenotype, thus losing their vision system through evolution has significant energy savings and thus improves survivability.  As such, we are inspired to computationally restrict resources available to descendant networks to encourage the evolution of highly-efficient deep neural networks.

Considering the aforementioned example, the descendant networks must take on network architectures with more efficient energy consumption than this original ancestor network to be able to survive.  The main factor in energy consumption is the quantity of synapses and neurons in the network. Therefore, to mimic environmental constraints that encourage the evolution of highly-efficient deep neural networks, we introduce an environmental constraint $\mathcal{F(\mathcal{E})}=C$ that probabilistically constrains the quantity of synapses that can be synthesized in the descendant network (which in effect also constrains the quantity of neurons that can be synthesized), such that descendant networks are forced to evolve more efficient network architectures than their ancestor networks.

Therefore, given $P(S_g|\mathcal{W}_{g-1})$ and $\mathcal{F(\mathcal{E})}=C$, the synthesis probability $P(\mathcal{H}_g)$ can be formulated as
\begin{align}
P(\mathcal{H}_g) = C \cdot P(S_g|\mathcal{W}_{g-1}),
\label{synthprob}
\end{align}
where $C$ is the highest percentage of synapses desired in the descendant network.  The random element of the network synthesis process mimics the random mutation process and promotes network architectural diversity.

\begin{figure*}[t]
{\center
\includegraphics[width = 16 cm]{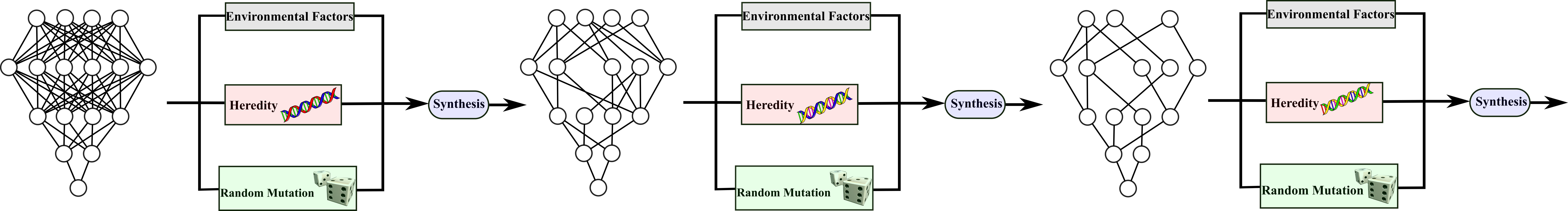}
\caption{Evolutionary synthesis process of highly-efficient deep neural networks.}
\label{Fig:EvoNet}}
\vspace{- 0.5 cm}
\end{figure*}

Given the probabilistic framework introduced above, the proposed evolutionary synthesis of highly-efficient deep neural networks can be described as follows (see Figure~\ref{Fig:EvoNet}).  Given an ancestor network at generation $g-1$, a synaptic probability model $P(S_{g}|\mathcal{W}_{g-1})$ is constructed according to Eq.~\ref{synaptiveprob}.  Using $P(S_{g}|\mathcal{W}_{g-1})$ and environmental constraint $\mathcal{F(\mathcal{E})}$, a synthesis probability $P(\mathcal{H}_g)$ is constructed according to Eq.~\ref{synthprob}.  To synthesize a descendant nework at generation $g$, each synapse $s_{g,k}$ in the descendant network is synthesized randomly as follows:
\begin{align}
s_{g,k}~{\rm exists~in~}~\mathcal{H}_g~{\rm if}~ P(s_{g,k}) \geq U(0;1),
\end{align}
\noindent where $U(0;1)$ is a uniformly distributed random number from a uniform distribution between 0 and 1.  The synthesized descendant networks at generation $g$ are then trained into fully-functional networks, like one would train a newborn, and the evolutionary synthesis process is repeated for producing successive generations of descendant networks.

\section{Experimental Results}

To investigate the efficacy of the proposed evolutionary synthesis of highly-efficient deep neural networks, experiments were performed using the MSRA-B~\cite{MSRAB} and HKU-IS datasets~\cite{li2015visual} for the task of visual saliency. This task was chosen given the importance for biological beings to detect objects of interest (e.g., prey, food, predators) for survival in complex visual environments, and can provide interesting insights into the evolution of networks.  Three generations of descendant deep neural networks (second, third, and fourth generations) were synthesized within an artificially constrained environment beyond the original, first-generation ancestor network.  The environmental constraint imposed during synthesis in this study is that the descendant networks should not have more than 40\% of the total number of synapses that its direct ancestor network possesses (i.e., $C=0.4$), thus encouraging the evolution of highly-efficient deep neural networks.  The network architecture of the original, first generation ancestor network used in this study, and details on the tested datasets and performance metrics are as follow.

\noindent \textbf{Network architecture.} The network architecture of the original, first generation ancestor network used in this study builds upon the VGG16 very deep convolutional neural network architecture~\cite{simonyan2014very} for the purpose of image segmentation as follows.  The outputs of the c3, c4, and c5 stacks from the VGG16 architecture are fed into newly added c6, c7, c8 stacks, respectively.  The output of the c7 and c8 stacks are then fed into d1 and d2 stacks.  The concatenated outputs of the c6, d1, and d2 stacks are then fed into the c9 stack. The output of the c5 stack is fed into c10 and c11 stacks. Finally, the combined output of the c9, c10 and c11 stacks are fed into a softmax layer to produce final segmentation result.  The details of  different stacks are as follows: c1: 2 convolutional layers of 64, $3 \times 3 $ local receptive fields, c2: 2 convolutional layers of 128, $3 \times 3$ local receptive fields,  c3: 3 convolutional layers of 256, $3 \times 3$ local receptive fields, c4: 3 convolutional layers of 512, $3 \times 3$ local receptive fields, c5: 3 convolutional layers of 512, $3 \times 3$ local receptive fields, c6: 1 convolutional layers of 256, $3 \times 3$ local receptive fields, c7 and c8: 1 convolutional layers of 512, $3 \times 3$ local receptive fields, c9: 1 convolutional layers of 384, $1 \times 1$ local receptive fields, c10 and c11: 2 convolutional layers of 512, $11 \times 11$ local receptive fields and  384, $1 \times 1$ local receptive fields, d1 and d2 are deconvolutional layers.

\begin{figure}[t]
	\setlength\tabcolsep{0.1 cm}
	\begin{center}
		\begin{tabular}{cccc}
			\includegraphics[width = 2 cm]{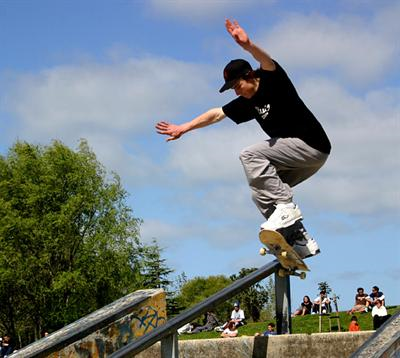}&
			\includegraphics[width = 2 cm]{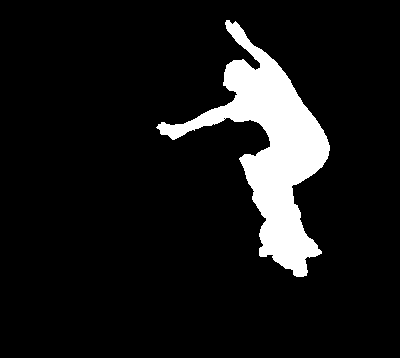}&
			\includegraphics[width = 2.1 cm]{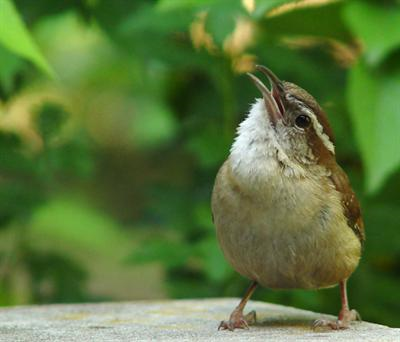}&
			\includegraphics[width = 2.1 cm]{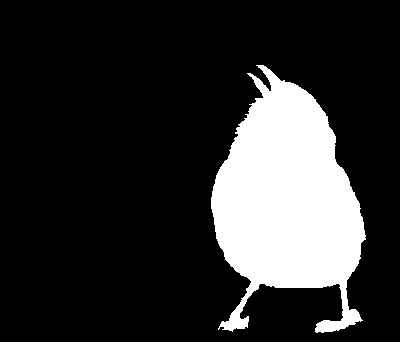}\\

			\includegraphics[width = 2 cm]{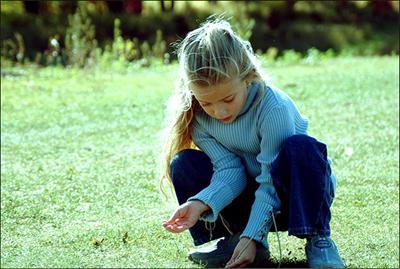}&
			\includegraphics[width = 2 cm]{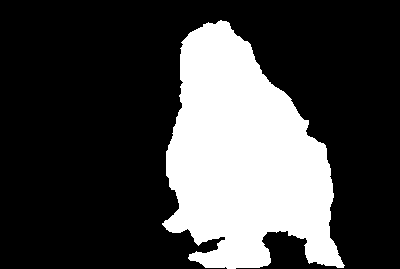}&
			\includegraphics[width = 2.05 cm]{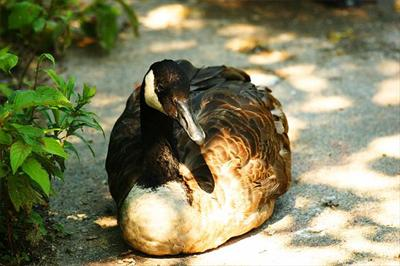}&
			\includegraphics[width = 2.05 cm]{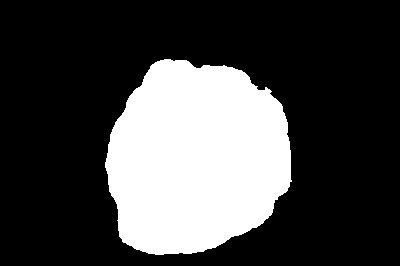}\\
			
			\includegraphics[width = 2 cm]{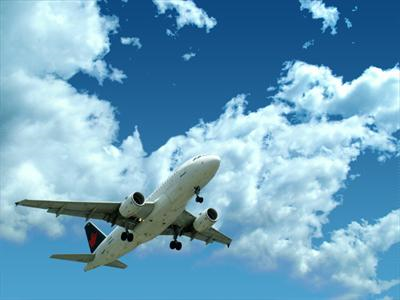}&
			\includegraphics[width = 2 cm]{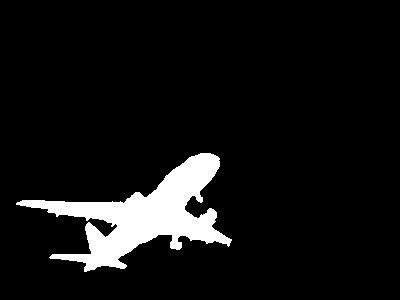}&
			\includegraphics[width = 2 cm]{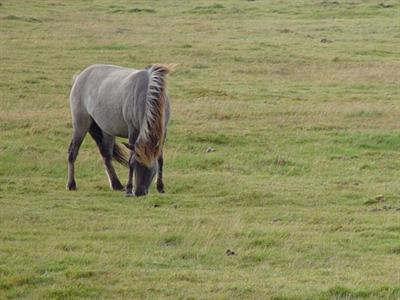}&
			\includegraphics[width = 2 cm]{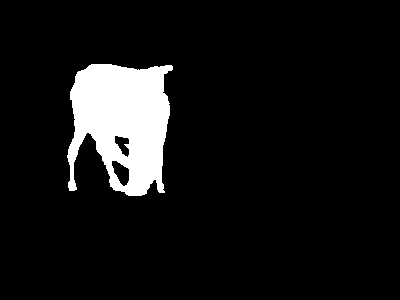}\\
			
		\end{tabular}
		\caption{MSRA-B image dataset:  This dataset contains 5000 natural images divided into 2500, 500 and 2000 images as training, validation and test samples, respectively. The ground truth maps are provided with pixel-wise annotation.  Examples of images and their corresponding ground truth maps in the MSRA-B image dataset are shown here.}
		\label{Fig:MSRA-B}
	\end{center}
\vspace{- 0.5 cm}
\end{figure}

\begin{figure}[t]
	\centering
	\setlength\tabcolsep{0.1 cm}
	\begin{center}
		\begin{tabular}{cccc}
			\includegraphics[width = 2 cm]{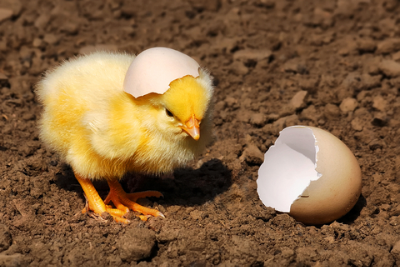}&
			\includegraphics[width = 2 cm]{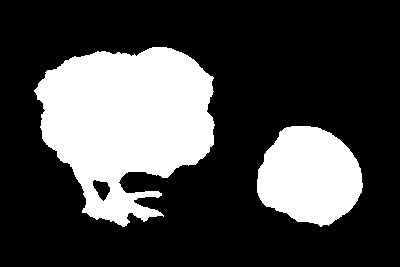}&
			\includegraphics[width = 2.1 cm]{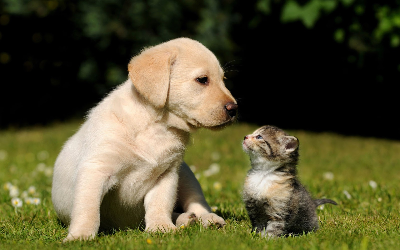}&
			\includegraphics[width = 2.1 cm]{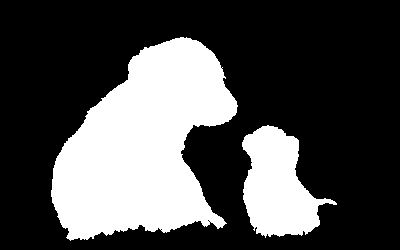}\\

			\includegraphics[width = 2 cm]{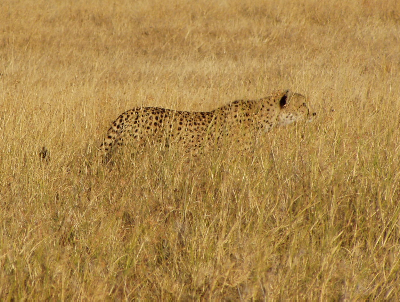}&
			\includegraphics[width = 2 cm]{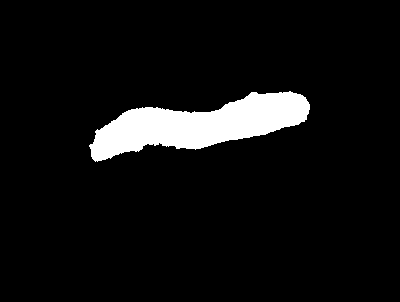}&
			\includegraphics[width = 2.05 cm]{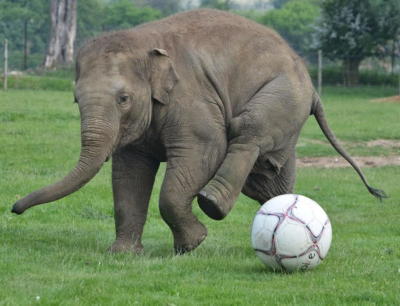}&
			\includegraphics[width = 2.05 cm]{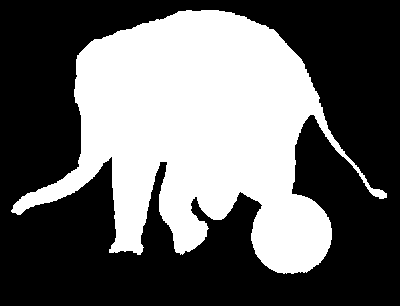}\\
			
			\includegraphics[width = 2 cm]{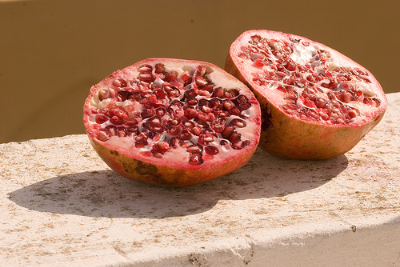}&
			\includegraphics[width = 2 cm]{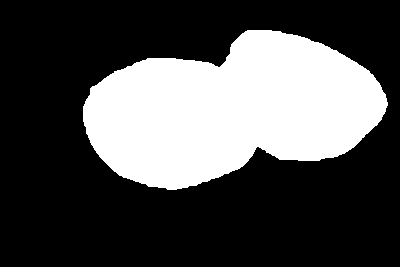}&
			\includegraphics[width = 2 cm]{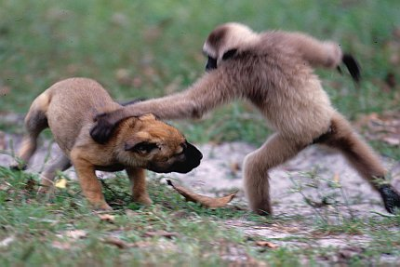}&
			\includegraphics[width = 2 cm]{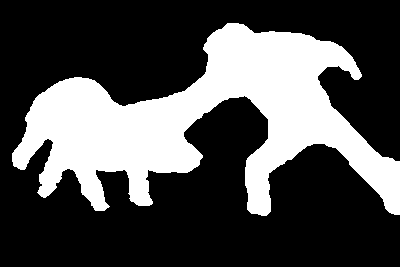}\\
			
		\end{tabular}
		\caption{HKU-IS image dataset:  This dataset contains 4447 natural images, and the entire dataset is used as a testing group for the descendant deep neural networks trained on the training group of the MSRA-B dataset.}
		\label{Fig:HKU-IS}
	\end{center}
\vspace{- 0.5 cm}
\end{figure}

\noindent \textbf{Datasets.} The MSRA-B dataset~\cite{MSRAB} consists of 5000 natural images and their corresponding ground truth maps where the salient objects in the images are segmented with pixel-wise annotation. The dataset is divided into training, validation and testing groups containing 2500, 500 and 2000 images, respectively.
 Figure~\ref{Fig:MSRA-B}
  The HKU-IS dataset~\cite{li2015visual} consists of 4447 natural images and their corresponding ground truth maps where the salient objects in the images are segmented with pixel-wise annotation. The entire dataset is used as a testing group for the descendant networks trained on the training group of the MSRA-B dataset.
   Figure~\ref{Fig:HKU-IS}
illustrates some of the example images from the dataset with their corresponding ground truths.

\noindent \textbf{Performance metrics.} To evaluate the performance of the evolved descendant deep neural networks at different generations, the MAE, F$_\beta$ score (where $\beta^2$=0.3~\cite{li2015visual}) metrics were computed for each of the descendant deep neural networks across the 2000 test images of the MSRA-B dataset that were not used for training.  As a reference, the same performance metrics was also computed for the original, first generation ancestor deep neural network.

\noindent \textbf{Architectural efficiency over successive generations.}
The detailed experimental results describing the number of synapses, architectural efficiency (defined here as the reduction of synapses in the network compared to the original, ancestor deep neural network in the first generation), F$_\beta$ score, and mean absolute error (MAE) presented in Table 1 and Table 2 for the MSRA-B and HKU-IS datasets, respectively.  A number of insightful observations can be made with respect to change in the architectural efficiency over successive generations of descendant deep neural networks.
\begin{table}[ht]
	\begin{center}
		\footnotesize
		\setlength\tabcolsep{0.1 cm}
		\caption{Performance metrics for different generations of synthesized offspring networks for MSRA-B dataset}
		\label{Tab:QRes}
		\begin{tabular}{l||cccccc}
			Generation  &  \scriptsize Number of synapses &\scriptsize Architectural efficiency &\scriptsize F$_\beta$ score  	 &\scriptsize MAE  \\ \hline \hline
			1 &63767232   & 1X &0.875   &0.0743  \\
			2 &15471797   &4.12X  & 0.876    &0.0739          \\
			3 &3603007   &17.69X &0.861     &0.0813    \\
			4 &1333010   &47.83X &0.850     &0.0863    \\
		\end{tabular}
	\end{center}
\vspace{- 0.5 cm}
\end{table}

First, it can be observed that the performance differences from one generation of descendant networks to the next generation are small for MSRA-B ($<$3\% between first generation and the fourth generation), while the performance differences are small for HKU-IS between the first two generations ($<$0.5\%) before larger performance differences in the third and fourth generations ($<$8\% between the first and fourth generations).  These results indicate that the modeling power of the ancestor network are well-preserved in the descendant networks.

Second, it can be observed that the descendant networks in the second and third generations can achieve state-of-the-art F$_\beta$ scores for MSRA-B (0.876 at second generation and 0.861 at third generation, compared to 0.865 as reported by Li et al.~\cite{li2015visual} for their state-of-the art visual saliency method), while having network architectures that are significantly more efficient compared to the first generation ancestor network (\textbf{$\sim$18-fold} decrease in synapses).  A similar trend was observed for HKU-IS, though persisting only in the second generation (0.826 compared to 0.8 reported in~\cite{li2015visual}, while achieving a \textbf{$\sim$4-fold} decrease in synapses over ancestor network).  What is more remarkable is that the descendant network at the fourth generation maintains strong F$_\beta$ scores (0.850 for MSRA-B and 0.753 for HKU-IS), while having network architectures that are incredibly efficient (\textbf{$\sim$48-fold} decrease in synapses) compared to the first generation ancestor network.  This \textbf{$\sim$48-fold} increase in architectural efficiency while maintaining modeling power clearly show the efficacy of producing highly-efficient deep neural networks over successive generations via the proposed evolutionary synthesis.

\begin{table}[ht]
	\begin{center}
		\footnotesize
		\setlength\tabcolsep{0.1 cm}
		\caption{Performance metrics for different generations of synthesized offspring networks for HKU-IS dataset}
		\label{Tab:QResHKUIS}
		\begin{tabular}{l||cccccc}
			\scriptsize Generation  &\scriptsize  Number of synapses &\scriptsize Architectural efficiency &\scriptsize F$_\beta$ score  	&\scriptsize MAE  \\ \hline \hline
			1 &63767232   & 1X &0.830   &0.0914  \\
			2 &15471797  &4.12X  & 0.826    &0.0911          \\
			3 &3603007  &17.69X &0.775    &0.1087    \\
			4 &1333010  &47.83X &0.753    &0.1190    \\
		\end{tabular}
	\end{center}
\end{table}

\begin{figure}[h]
	\centering
	\setlength\tabcolsep{0.05 cm}
	\begin{center}
		\begin{tabular}{ccccc}
			\includegraphics[width = 1.7 cm]{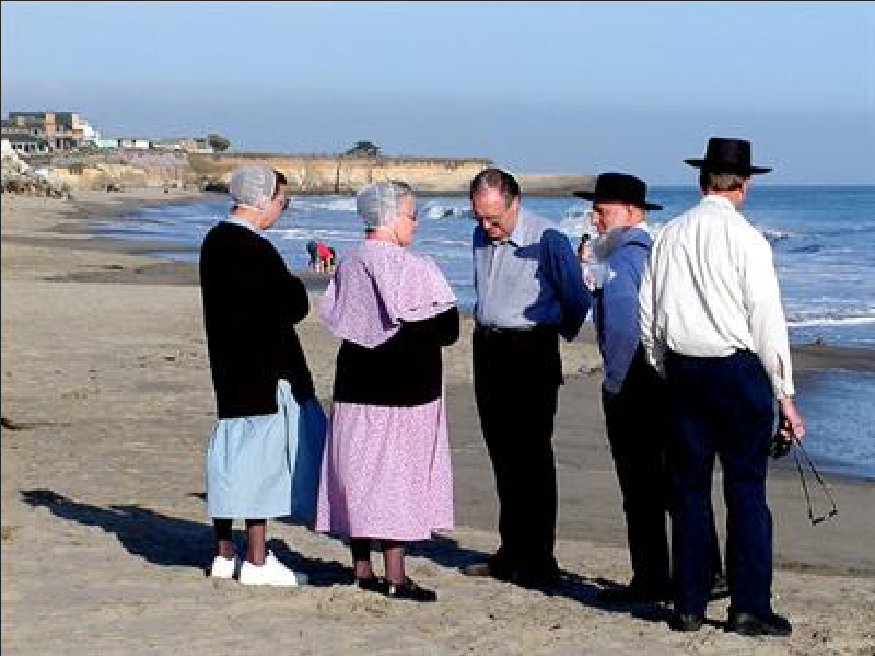}&
			\includegraphics[width = 1.7 cm]{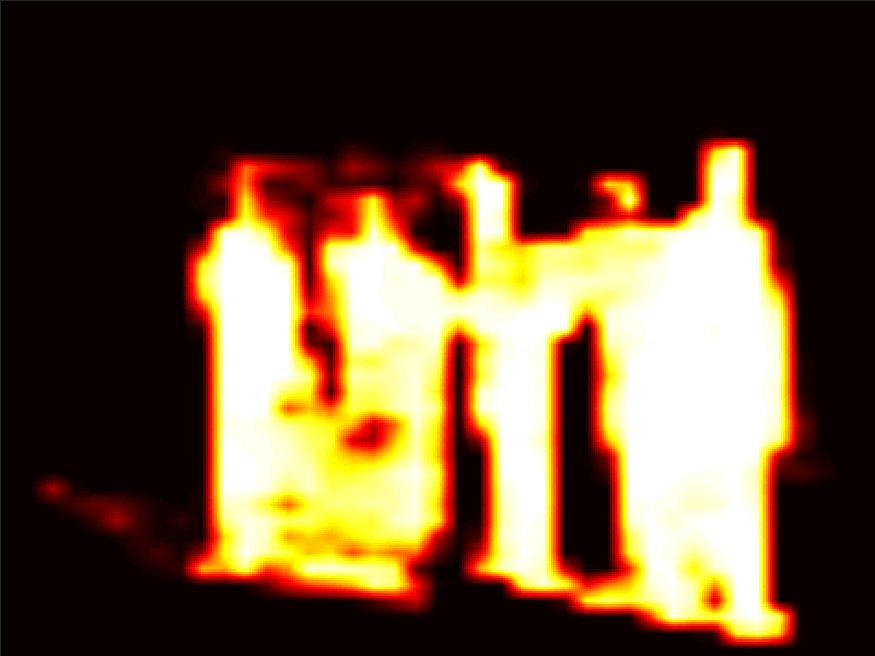}&
			\includegraphics[width = 1.7 cm]{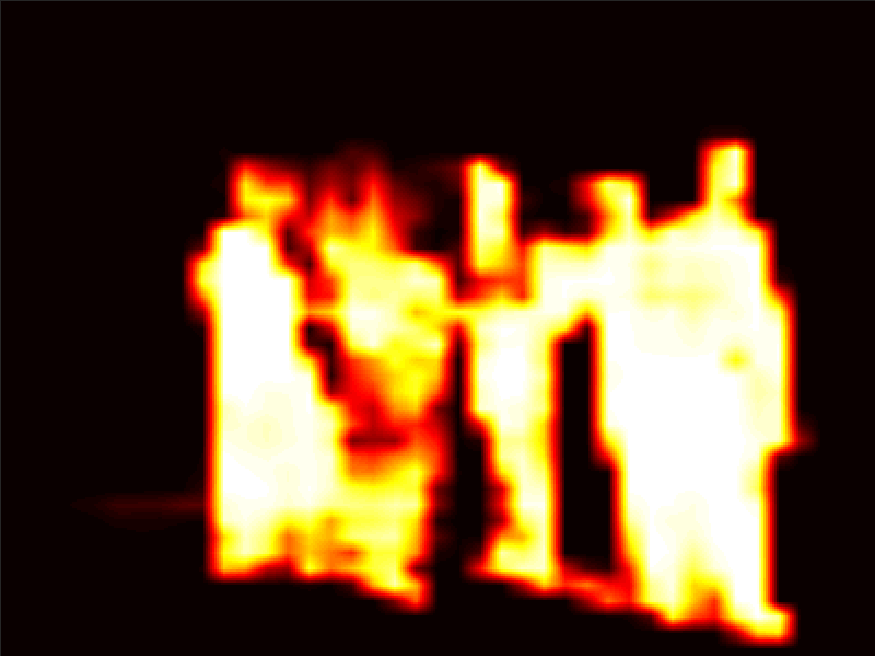}&
			\includegraphics[width = 1.7 cm]{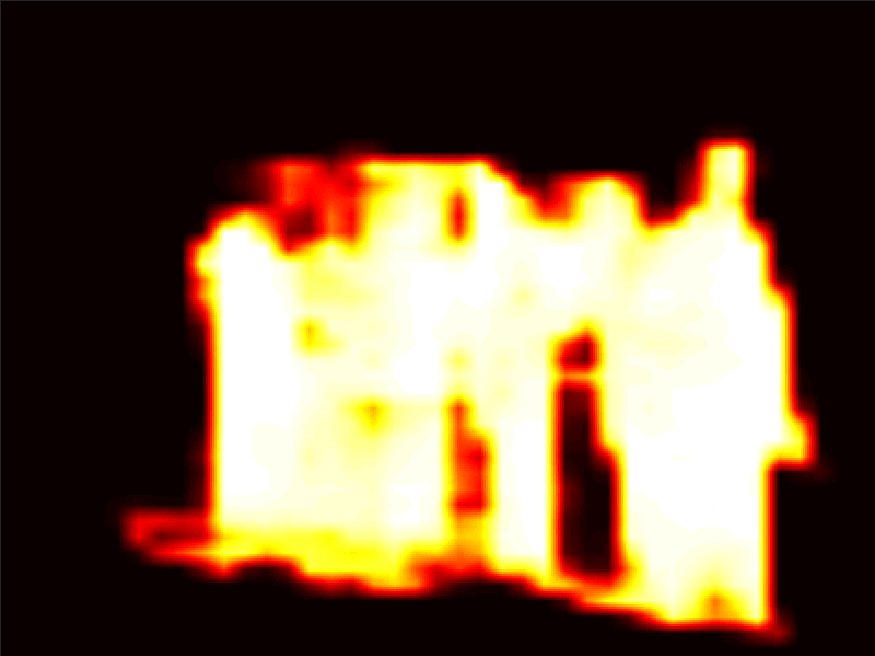}&
			\includegraphics[width = 1.7 cm]{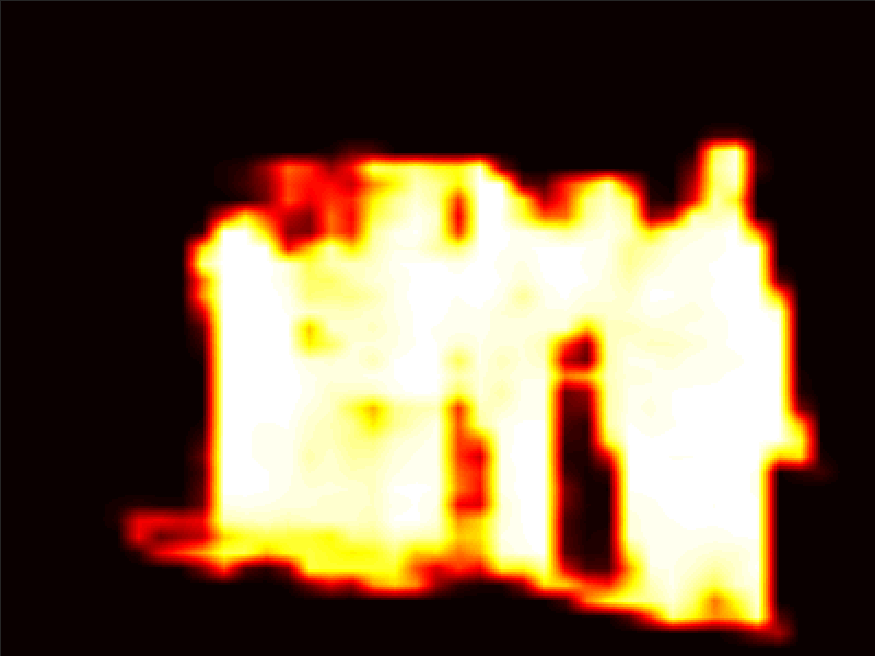}\\
		
			\includegraphics[width = 1.7 cm]{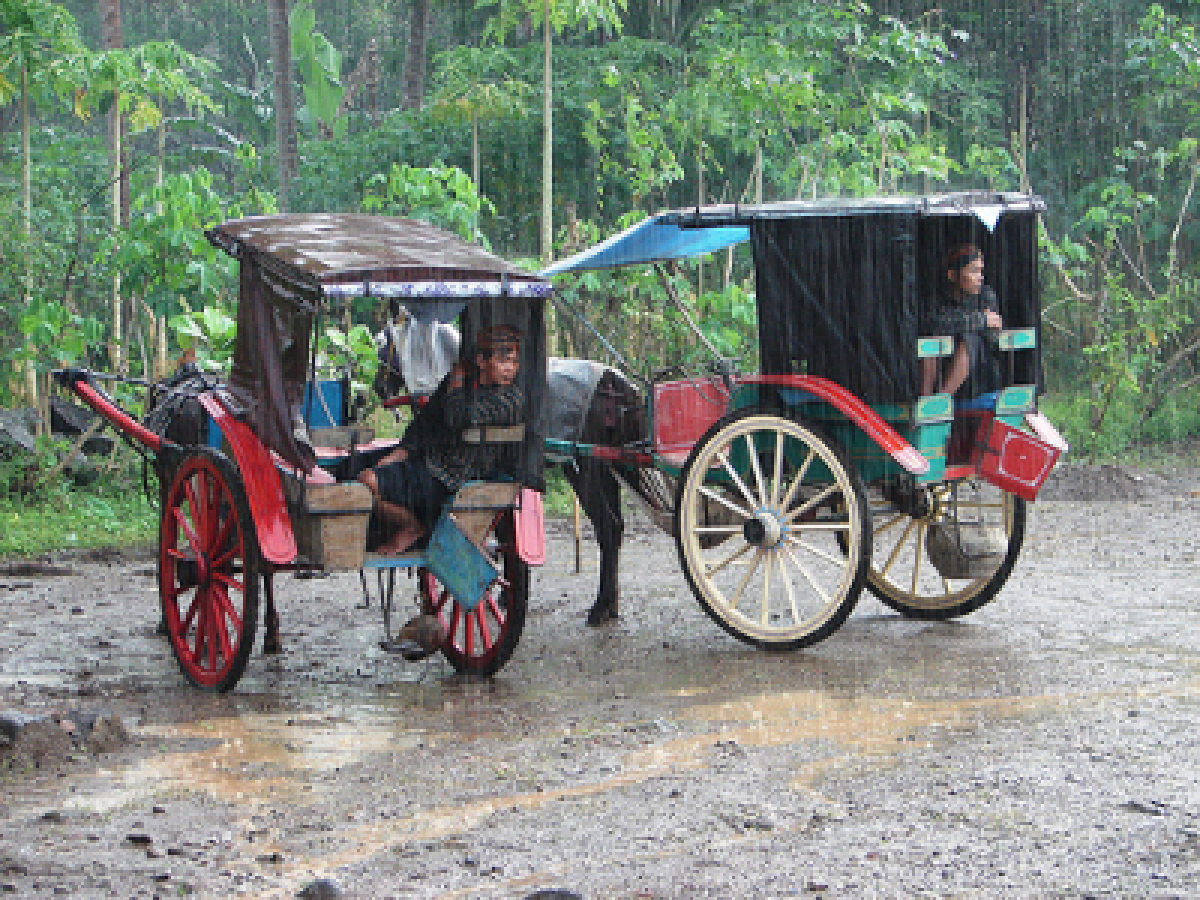}&
			\includegraphics[width = 1.7 cm]{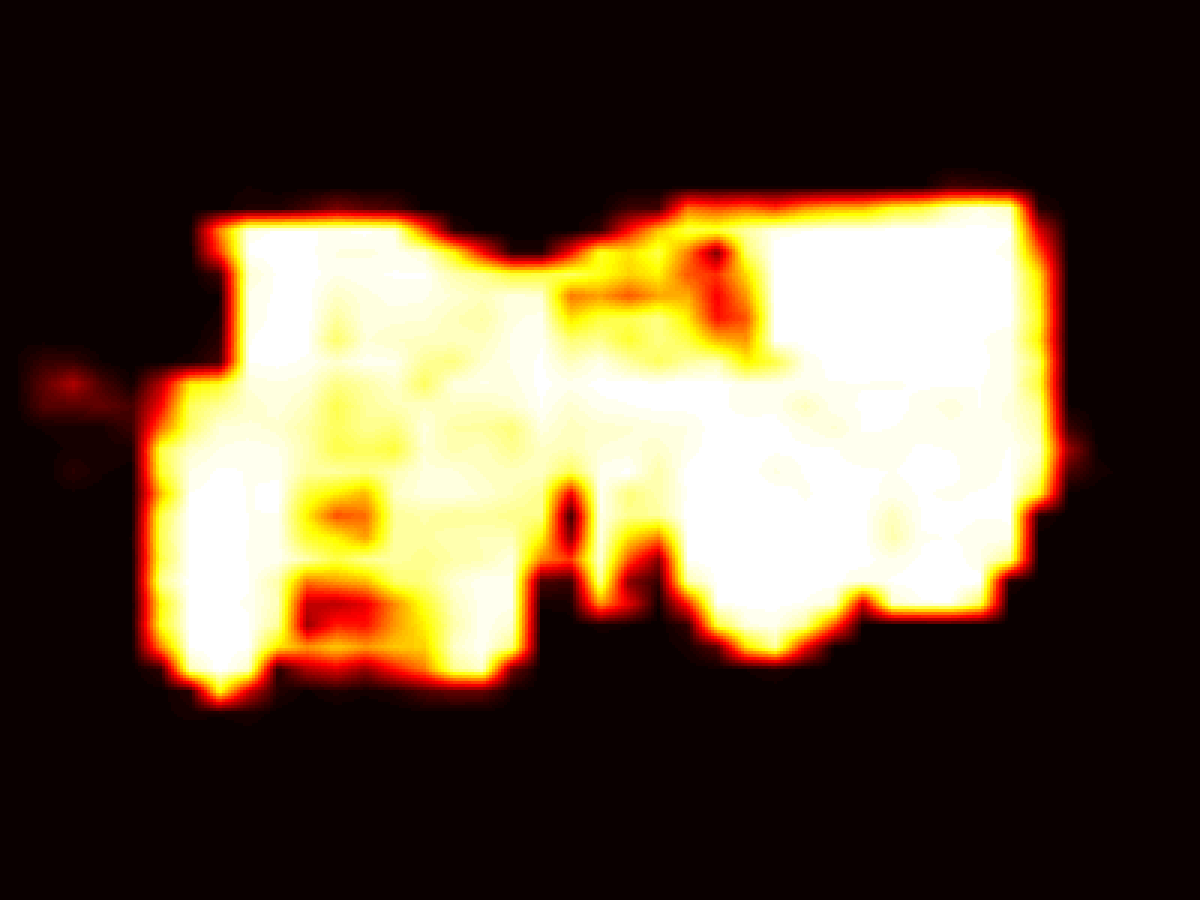}&
			\includegraphics[width = 1.7 cm]{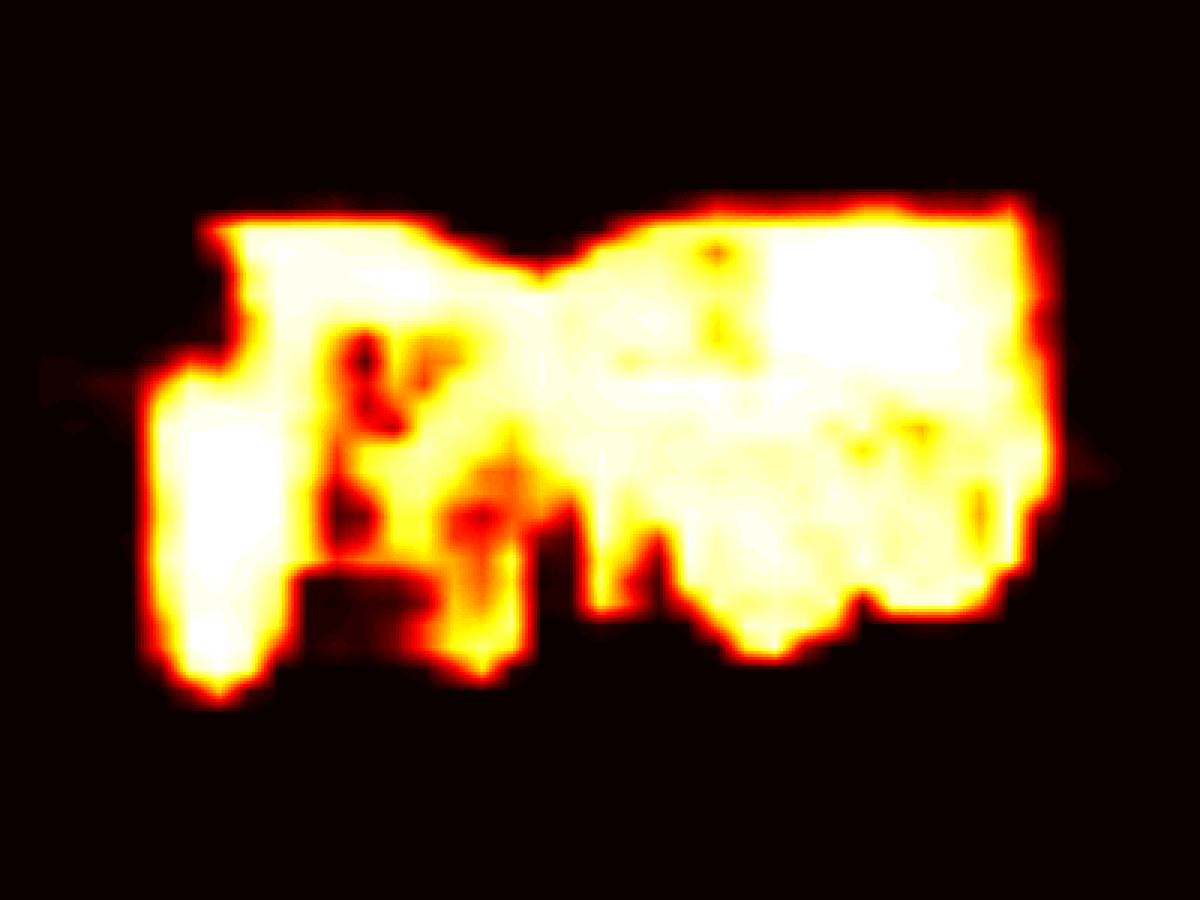}&
			\includegraphics[width = 1.7 cm]{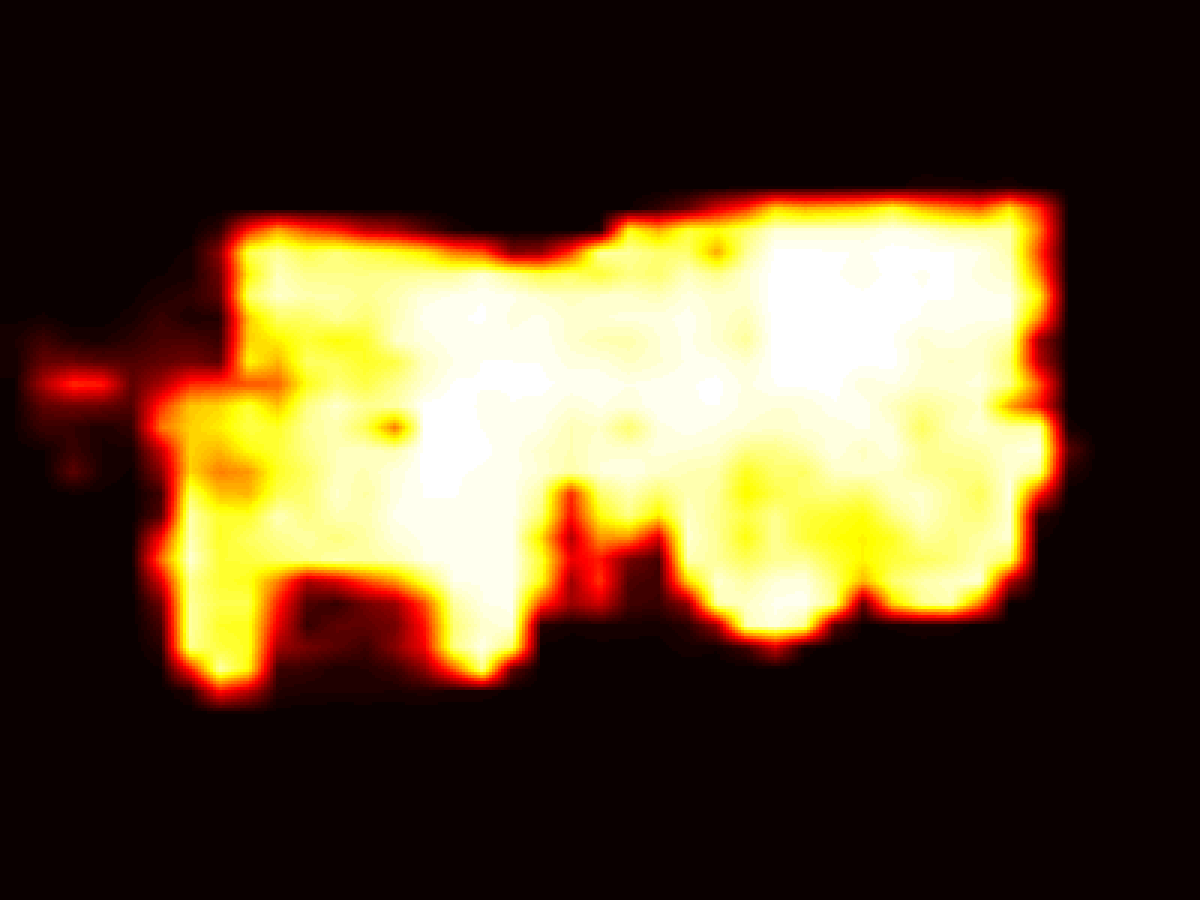}&
			\includegraphics[width = 1.7 cm]{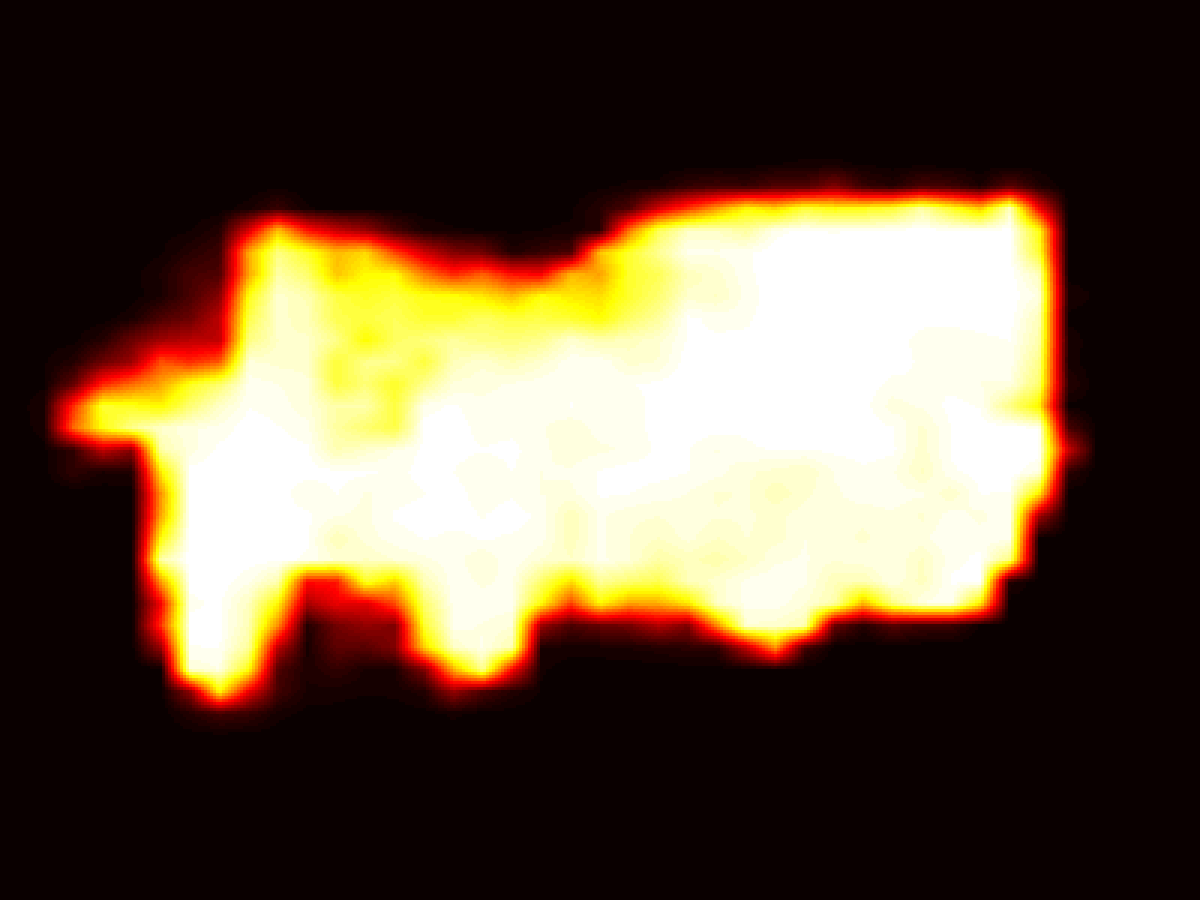}\\
			\scriptsize Image &\scriptsize Generation 1 &\scriptsize Generation 2 &\scriptsize Generation 3 &\scriptsize Generation 4
			
		\end{tabular}
		\caption{Example test images from the tested datasets, and the corresponding visual saliency maps generated by the descendant deep neural networks at different generations.}
		\label{Fig:saliency}
	\end{center}
\end{figure}

\noindent \textbf{Visual saliency variations over successive generations.}
To gain additional insights, Figure~\ref{Fig:saliency}
demonstrate example test images from the MSRA-B dataset and the HKU-IS dataset, respectively, along with the corresponding visual saliency maps generated by the descendant networks at different generations.  It can be observed that the descendant networks at all generations consistently identified the objects of interest in the scene as visually salient.  It is also interesting to observe that by the fourth generation, with a $\sim$48-fold decrease in synapses compared to the first generation ancestor network, the ability to distinguish fine-grained visual saliency starts to diminish.  These observations are interesting in that, similar to biological evolution, they show that the descendant networks evolved over successive generations in such a way that important traits (e.g., general ability to identify salient objects) are retained from its ancestors while less important traits (e.g., ability to distinguish fine-grained saliency) diminish in favor of adapting to environmental constraints (e.g., growing highly-efficient architectures due to imposed constraints).

These experimental results show that, by taking inspiration from biological evolution, the proposed evolutionary synthesis of deep neural networks can lead to the natural evolution of deep neural networks over successive generations into highly efficient, yet powerful deep neural networks, and thus a promising direction for future exploration in deep learning.

\vspace{- 0.5 cm}

\section*{Author Contributions}
A.W. conceived the concept of evolutionary synthesis for deep learning proposed in this study. M.S. and A.W. formulated the evolutionary synthesis process proposed in this study. A.M. implemented the evolutionary synthesis process and performed all experiments in this study.  A.W., M.S., and A.M. all participated in writing this paper.

\renewcommand{\refname}{\normalfont\selectfont\normalsize\bf References}
\small{
	\bibliographystyle{IEEEtran}
	\bibliography{refs}
}







\end{document}